\documentclass{article}

\usepackage{arxiv}

\usepackage[utf8]{inputenc} 
\usepackage[T1]{fontenc}    
\usepackage{hyperref}       
\usepackage{url}            
\usepackage{booktabs}       
\usepackage{amsfonts}       
\usepackage{nicefrac}       
\usepackage{microtype}      
\usepackage{lipsum}		    
\usepackage{graphicx}
\usepackage{doi}

\usepackage{amsmath}
\usepackage{mathtools}
\usepackage{float}
\usepackage{xcolor}
\usepackage{array}
\graphicspath{{./}}

\title{CogSense: A Cognitively Inspired Framework for Perception Adaptation}

\author{Hyukseong Kwon \\ hkwon@hrl.com \and Amir Rahimi \\ amrahimi@hrl.com \and Kevin G. Lee \\ klee.ven@hrl.com \and Amit Agarwal \\ aagarwal@hrl.com \and Rajan Bhattacharyya \\ rbhattac@hrl.com}

\date{HRL Laboratories, LLC., 3011 Malibu Canyon Road, Malibu, CA 90265, USA}

\renewcommand{\undertitle}{}

\hypersetup{
pdftitle={A template for the arxiv style},
pdfsubject={q-bio.NC, q-bio.QM},
pdfauthor={Hyukseong ~Kwon, Amir ~Rahimi, Kevin G. ~Lee, Amit ~Agarwal, Rajan ~BBhattacharyya},
pdfkeywords={Perception error, Temporal logic, Perception adaptation},
}

\begin{document}
\maketitle

\begin{abstract}
This paper proposes the CogSense system, which is inspired by sense-making cognition and perception in the mammalian brain to perform perception error detection and perception parameter adaptation using probabilistic signal temporal logic. As a specific application, a contrast-based perception adaption method is presented and validated. The proposed method evaluates perception errors using heterogeneous probe functions computed from the detected objects and subsequently solves a contrast optimization problem to correct perception errors. The CogSense probe functions utilize the characteristics of geometry, dynamics, and detected blob image quality of the objects to develop axioms in a probabilistic signal temporal logic framework. By evaluating these axioms, we can formally verify whether the detections are valid or erroneous. Further, using the CogSense axioms, we generate the probabilistic signal temporal logic-based constraints to finally solve the contrast-based optimization problem to reduce false positives and false negatives.  
\end{abstract}

\keywords{Perception error \and Temporal logic \and Perception adaptation}

\section{Introduction}
\label{sec:intro} 

Perception errors are still challenging issues despite performance improvements in perception systems in the last decade. In autonomous driving or navigation systems, incorrect detections threaten the safe and robust performance of fully autonomous systems. Most modern perception systems utilize a pipeline of algorithms to combine functions of object detection and tracking, however they perform their inference in a purely feedforward manner. Perceptual processing in the mammalian brain, however, has significant feedback, as evidenced in the presence of dense connectivity from higher-order visual areas encoding object categories to early visible areas encoding features \cite{c00}. This principle of feedback extends to intelligence after perception, in \textit{cognition}, where the brain moves far beyond simply binding visual features and matching to the most likely feature distribution to recognize an object, but instead engages in cycles of attention, hypothesis revision, and decision making about conflicting features to infer the true identity of an object under uncertain conditions. The cognitive capability of the human brain to make inferences under uncertainty by foraging, adaptation, evaluation, and decision making is known as sense-making and has been studied using behavioral experiments and neurocognitive models \cite{c17, c18, c19}. There have also been approaches attempting to embed this feedback into deep neural networks \cite{c28,c29,c30}.

Evidence from neuroimaging illustrates the rich top-down feedback connections that provide the platform for attentive perceptual cognition in the brain.  These connections extend from areas that perform decision-making in the prefrontal cortex back to modulate activity in the sensory cortical areas that provide the visual features themselves \cite{c20, c21}. Further, diverse cortical areas that process the semantic properties of objects feed activity to the heteromodal association areas that bind these semantic properties with their sensory properties (including visual, auditory, tactile features) to combine perception and semantics in cognition \cite{c22, c23}. 

Adaptation in the brain occurs in many ways under a multitude of mechanisms, from local to global. For instance, when we receive feedback that demonstrates significant surprise, or unexpected uncertainty of our model of the world, the brain adapts its processing for a remarkable ecological endeavor: controlling attention to acquiring new statistics of the world around it to make better inferences by adapting at multiple levels through a neuromodulatory cascade of norepinephrine \cite{c25}. At the decision-making level, neurobiological circuits that maintain the inference model are directed to release the current model from working memory in the prefrontal cortex. Feedback to the sensor occurs at the sensory level: the pupils in the eye dilate to acquire more stimulus. In turn, experimental evidence also supports mid-level cognitive adaptation in the level of alertness and orienting behavior to reacquire new statistics of the world \cite{c24}. Indeed, such models have been used in artificial neural networks for adapting behavior in robots \cite{c26, c27}. 

To this end, we are inspired by sense-making cognition as a model for intelligence in perception, using feedback, adaptation, and semantics. To enable fully autonomous systems, sense-making cognition for intelligent perception is necessary under uncertain conditions, including corner cases. In this paper, we demonstrated an approach to intelligent perception in a system that detects perception errors, and uses the principles of feedback and adaptation to correct them.

\section{Related Work}
\label{sec:related}

In the context of robotics, Perception systems need to have strong confidence about their environment. Perception errors can lead to disastrous consequences for the robot or its surroundings, for example, autonomous car crashes. Today, Deep Neural Networks (DNN) are commonly used for perception; however, they are notoriously easy to fool \cite{c41} and do not output calibrated confidences \cite{c31}. To prevent this vulnerability, researchers have investigated a more global measure called uncertainty. There are two kinds of uncertainty: aleatoric and epistemic. Aleatoric uncertainty is the uncertainty of our input, and epistemic uncertainty is the uncertainty of our model. Some methods to estimate epistemic uncertainty are Bayesian Neural Networks \cite{c36}, ensemble methods \cite{c32,c33}, Monte Carlo dropout \cite{c34}, sampling-free methods \cite{c35}, or directly from the input \cite{c37}. Uncertainty can be used as an indicator of the likelihood of an error. If the uncertainty associated with output is high, the network is not very confident in its output, and that an error is likely. However, most of these methods do not study how to reduce the uncertainty associated with an input to improve performance. 

Other approaches to detect perception errors and fix them have focused on formally verifying the systems using temporal logic \cite{c1,c2,c3,c4}. This approach has the advantage of mathematically proven performance guarantees, an essential aspect of autonomous systems operating under uncertain conditions. However, most of these systems apply motion control to the autonomous platforms themselves rather than fix the perception systems. Unlike previous works, our work is the first to apply temporal logic to directly modulate the image contrast to the best of our knowledge. 

More direct methods to predict and fix errors use sensor cues from LIDAR or additional cameras to detect errors \cite{c38,c39}. In contrast to these works, our method requires no additional sensors besides a single camera and can be applied on top of any off-the-shelf object detector. 

Our approach builds upon contrast enhancement techniques. Early work in the perceptual domain has explored improving upon conventional contrast enhancement techniques by using adaptive histogram equalization \cite{c5}. Subsequent work has applied contrast enhancement to object detection, where it helps to accentuate features to detect objects more robustly \cite{c6,c7}. However, corresponding conventional methods use the image contrast information of the entire image. So, if some non-object areas cause high contrast, contrast adaptation cannot improve object detection. 

In our proposed method, we draw upon temporal logic-based methods and contrast enhancement techniques. We improve the perception system using feedback control of a contrast parameter only from detected objects within a formally verified system. 

The rest of the paper will be presented as follows:  In the next section, we describe the proposed perception error detection/correction system.  In Section \ref{sec:probes}, the probes converted from the perception data are explained. Then the detailed approach for error correction is presented in Section \ref{sec:ErrorCorrection}.  With that mechanism, Section  \ref{sec:ContrastAdaptation} goes into the details of the contrast-based perception adaptation.   And Section \ref{sec:experiments} validates the proposed methods, and finally Section \ref{sec:conclusion} concludes the paper.

\section{Overview}
\label{sec:overview} 

\begin{figure*}
  \centering
    \includegraphics[width=6in]{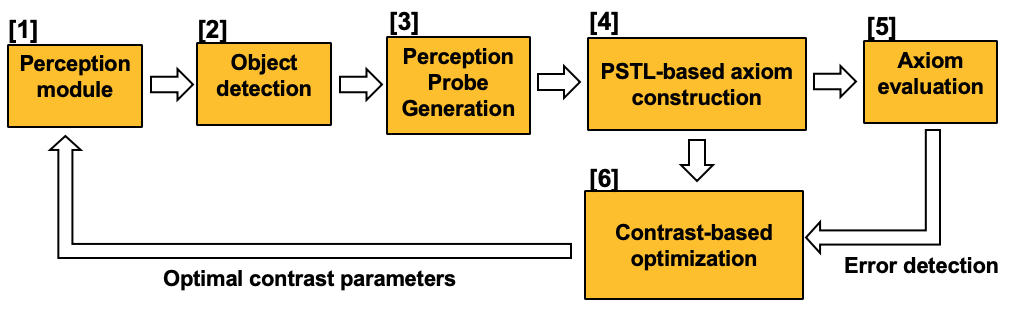}
  \caption{Overview architecture. We perform object detection on an input image and generate probes containing the statistics of the detections. Next we construct PTSL axioms and adapt the contrast of the input image by optimizing on the PTSL axioms. Finally we perform object detection again on the contrast-adjusted image.}
  \label{fig:overview}
\end{figure*}

The overall structure of the CogSense error detection and perception adaptation system is shown in Figure \ref{fig:overview}. Here, we review the construction of the CogSense system, from training data to characterizing the operational ranges of the perception system. First, from the perception data, we generate probes that describe detections' characteristics, such as the size of objects, their type, tracking deviations from a motion tracker, and image statistics such as contrast and entropy. Using the probes, we set up ‘Probabilistic Signal Temporal Logic (PSTL) \cite{c4},’ and PSTL provides axioms, each of which is constructed with a single or multiple probes with the corresponding statistical analyses.  As an intermediate process, those axioms provide the error analysis on detections. This is the cognitively inspired process described earlier, in which heterogeneous sources of information from multiple modalities, including physical, semantic, and image based statistics provide checks and balances on the plausibility of output detections from the perception system that are orthogonal to the measures of uncertainty described above. Finally, with these axiom-based constraints, we solve an optimization problem to synthesize controls for the perception modules to reduce perception errors and improve valid detection rates. This paper first mathematically describes the CogSense system for the heterogeneous probes for error detection using PSTL and then the PSTL-based optimization framework for controlling parameters to correct perception errors. The following section describes our experimental results that utilize heterogeneous probes in CogSense to detect errors and an image contrast optimization for perception adaptation to correct errors.

In more detail, our approach is as follows:

\begin{enumerate}
    \item Through the perception module, we receive input images of the scene.   
    \item Objects in the image are detected and recognized.  
    \item This module converts the perception data into probes that we use for signal temporal logic. 
    \item Probes are converted into the axioms under the probabilistic signal temporal logic structure.  
    \item The axioms are evaluated to verify if the corresponding observations are valid or erroneous based on the constraints using the statistically analyzed probe bounds.  
    \item If the axioms are invalid within certain probabilities, estimate the optimal contrast bound and entropy bound as perception module parameters to apply by solving the image contrast/entropy based optimization problem.   Finally, this estimated parameters are delivered back to the perception module to adjust its contrast and entropy. 
\end{enumerate}

\section{Perception probes generation and error evaluation}
\label{sec:probes} 

\begin{figure}[ht]
  \centering
    \includegraphics[width=5in]{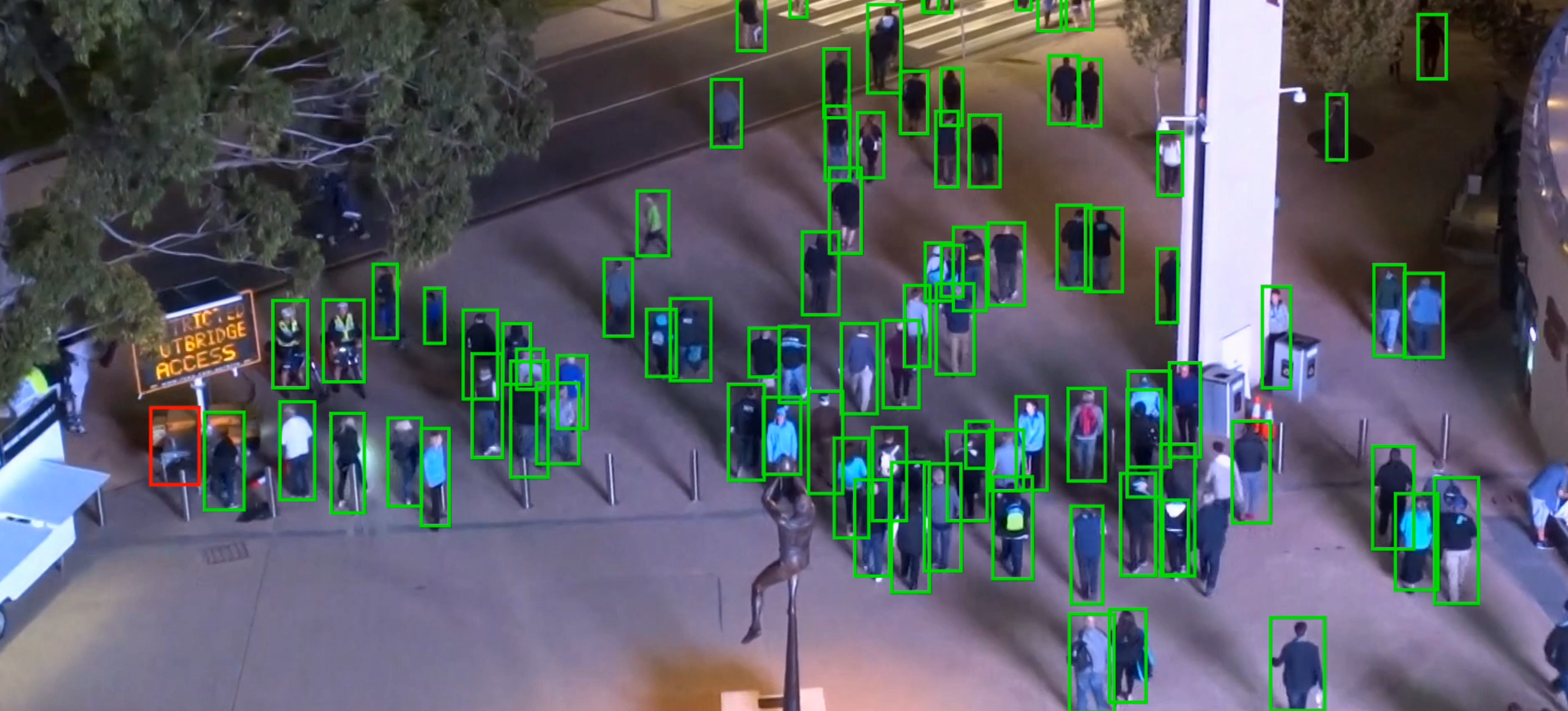}
  \caption{Sample image of person detection using YOLOv3}
  \label{fig:SampleYolo}
\end{figure}

The first step in the process is to obtain the perception data along with characteristics of detections and recognitions.  To get different types of characteristics efficiently, we used YOLOv3 \cite{c9} as shown in Figure \ref{fig:SampleYolo}. 

We can extract image-based bounding box information from detected object boxes in a single frame and the corresponding tracking information along with the image sequences.   We call this information "probes."  In our system, we have multiple probes such as detected object sizes, aspect ratios, recognition ID consistency, tracking deviations, and so on.  The following are the sample probes we use in this paper.   

\begin{itemize}
	\item Object size (in the image plane and in the world coordinate frame)
	\item Aspect ratio of the detected objects 
	\item Localization and tracking performance 
	\item Recognition confidence 
	\item Contrast of the detected boxes
	\item Entropy of the detected boxes
\end{itemize}

Each probe will be used to build constraints for perception adaptation using the "probabilistic signal temporal logic (PSTL) \cite{c4}."  PSTL uses probabilistic and temporal aspects on real-valued signals.   Let us assume that the signal $\xi$ at time $t$ satisfies a probabilistic atomic predicate $\lambda$.  If we briefly describe its formula, 

\begin{equation}
    (\xi, t) \models \lambda_{\alpha_{t}}^{\epsilon_{t}} \iff P(\lambda_{\alpha_{t}}(\xi(t))<0) > 1-\epsilon_{t}.  
    \label{eq:pstl_sample}
\end{equation}

In this predicate, $\lambda_{\alpha_{t}}^{\epsilon_{t}}$, $\alpha_t$ is a time-varying random variable and $1-\epsilon_{t}$ is the tolerance level in satisfying the probabilistic properties.  With this formula, we convert our probes into axioms.   

\begin{figure}[ht]
  \centering
    \includegraphics[width=5in]{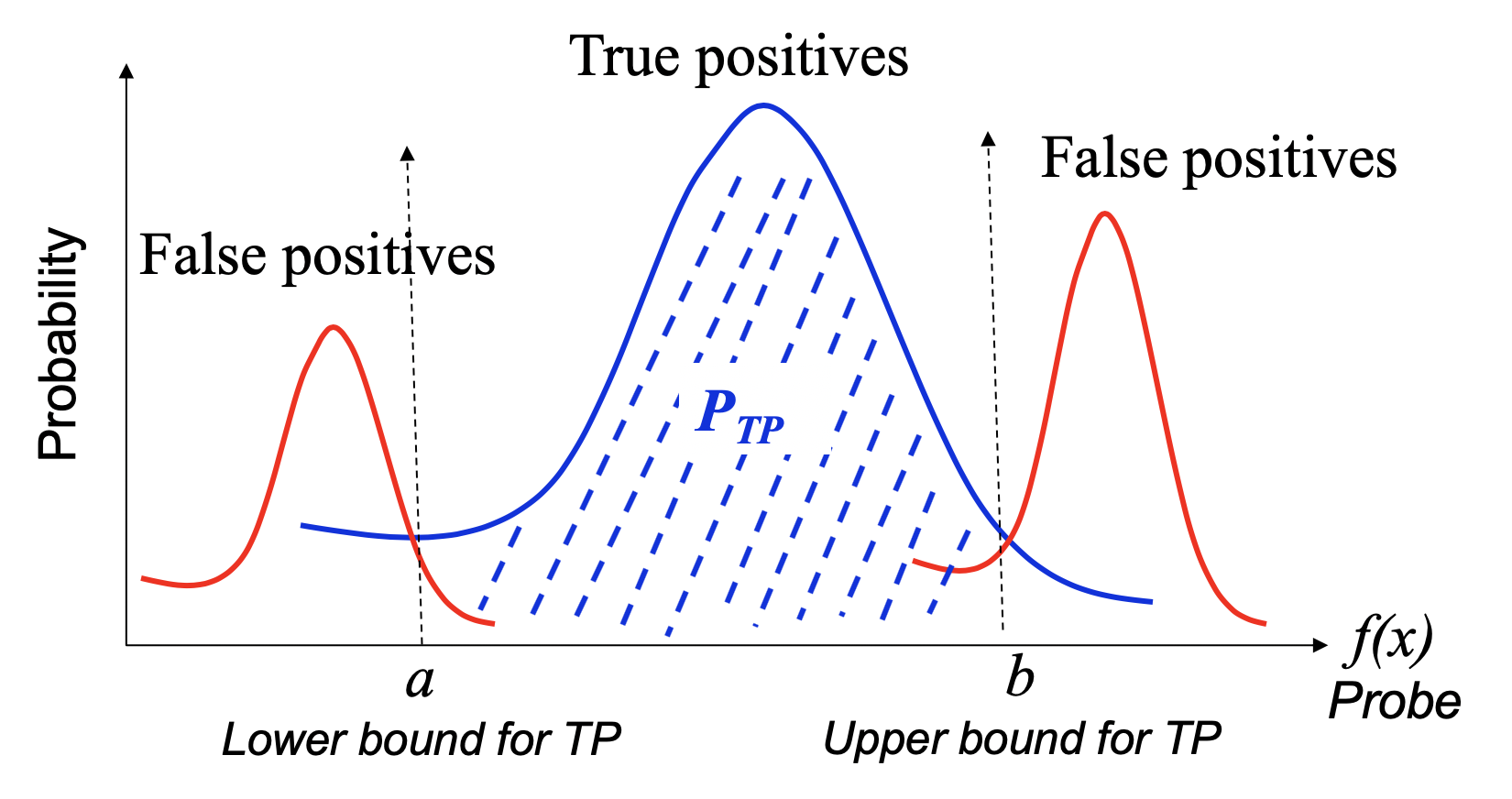}
  \caption{Example of probability distribution of a probe. There is some domain [a,b] where the probability of true positives is optimized.}
  \label{fig:ProbeStatistics}
\end{figure}

All of our detections are divided into true positives and false positives.  From the true positive and false positive detections, we can perform statistical analysis for each probe.  Figure \ref{fig:ProbeStatistics} shows a descriptive example of a probe.   For a detected object, $x$, we assume that we generate a probe, $f(x)$.  By analyzing the values from true positives and also those from false positives, we can obtain probabilistic distributions of true positives and false positives as shown in the figure.  We can define upper and lower bounds for true positives from the intersections between two different distribution graphs.  And the shaded area presents the confidence probability, $P_{TP}$, of the probe.  

If we describe this relation in a mathematical form (axiom) with the probabilistic inequality from the probabilistic signal temporal logic, it becomes as follows:

\begin{equation}
    \forall x, Pr(a \leq f(x,t_s:t_e) \leq b \rightarrow y) \geq P_{TP} 
    \label{eq:axiom_gen}
\end{equation}

where $Pr(\cdot)$ is the predicate and y is the true detection or recognition.  And $t_s:t_e$ means the time sequence between $t_s$ and $t_e$, so $f(x,t_s:t_e)$ is the probe sequence in the time frame of $t_s:t_e$.

Depending on the probe dimensions, the probabilistic function can also be multi-dimensional. By integrating all the available axioms from $x$, we can have a multi-dimensional range of the corresponding detection or recognition.   When a new probe violates the corresponding axioms more than a certain probabilistic threshold, we can verify that the corresponding detection is considered as erroneous, and the probability of an error is higher.   If it is categorized as erroneous, we apply the perception adaptation as shown in the next section.

\section{Perception error correction using the PSTL-constraint-based optimization}
\label{sec:ErrorCorrection} 

Detecting perception errors is not sufficient to recover the perception quality in the following image sequences.  Therefore, we also want to adjust perception modules to have more accurate and robust detections with that knowledge. In this paper, we propose a new optimization technique using the PSTL-constraint-based optimization with the following format: 

\begin{equation}
    u_{opt} = \arg\min_{u_t} J(x_t, u_t) 
    \label{eq:optimization}
\end{equation}

where $x_t$ is the probe state at time $t$ and $u_t$ is the control input to the perception module.  And $J(\cdot)$ is the cost function of estimating perception errors.  Our goal is to achieve the optimal $u_t$ to reduce perception errors.    Therefore, minimizing $J(\cdot)$ can achieve the optimal perception module control input.  Eventually, the final optimization formula with the two or more PSTL-based constraints for probes, $f(x_t)$, $g(z_t)$, etc. becomes, 

\begin{equation}
\begin{aligned} 
\begin{gathered}
    u_{opt} = \arg\min_{u_t} J(x_t, z_t, u_t) \\
    s.t. \\
    \forall x_t, Pr(a \leq f(x_t, t_s:t_e) \leq b \rightarrow y ) \geq P_{TP_{x}} \\
    \forall z_t, Pr(g(x_t, t_s:t_e) \leq c \rightarrow w ) \geq P_{TP_{z}} \\ 
    \cdots 
\end{gathered}
\end{aligned}
\label{eq:optimization_complete}
\end{equation}

where $y$ and $w$ are the true positive labels for the probes, $x_t$ and $x_t$, respectively.  $a$, $b$, and $c$ are probabilistic error bounds acquired from the true positive / false positive distribution described in Section \ref{sec:ErrorCorrection}.  And $P_{TP_{x}}$ and $P_{TP_{z}}$ are the probability thresholds for $x_t$ and $x_t$, respectively.

\section{Contrast-based perception adaptation }
\label{sec:ContrastAdaptation} 

To achieve the contrast-based perception adaptation, we first set up the object detection constraints using five different types of constraints: (1) Detection ID consistency (tracking of the same object); (2) Localization consistency within the expected trajectory; (3) Bounding box size consistency in the image plane; (4) Contrast matching in the desired range; and (5) Entropy matching in the desired range. Details for each constraint are presented below.     $t_k$ is the current time and $t_{(k-M)}$ is the time that the temporal logic window starts. 

\begin{itemize}

    \setlength{\itemsep}{1.2\baselineskip}

	\item Consistent detection 
	\begin{equation}
	    Pr({\sum_{t=t_k-t_{k-M}}^{t_k} det_{ID=X}}) > P_X
	\end{equation}

    where $P_X$ is the probabilistic threshold for consistent ID detections.  In this constraint, the detection ID is checked to be consistent.  If the IDs keep changing, the detection process cannot be robust.  
 
 	\item Bounding box size deviation over time 
    \begin{equation}
        Pr(|BB_D - BB_t|_{t=t_k-t_{k-M}}^{t_k}) < P_{BB}
    \end{equation}
    where $BB_t$ is the bounding box size at time $t$ and $BB_D$ is the desired bounding box size from its history.  And $P_{BB}$ is the probabilistic threshold for consistent bounding box size.  Usually, highly varying bounding box sizes for the same object (unless it does not approach or move away abruptly) indicates unreliable bounding box estimation (e.g. too sensitive with respect to lighting condition changes).  
 
	\item Localization deviation from the desired tracking trajectory
	\begin{equation}
	    Pr(|Path_{Desired} - loc_t|_{t=t_k-t_{k-M}}^{t_k}) < P_{loc}
	\end{equation}

    where $loc_t$ is the detected object’s location at time $t$ and $Path_{Desired}$ is its expected path from the history, and $P_loc$ is the probabilistic threshold for consistent localization.  Localization deviation usually comes from unreliable bounding box estimation, which indicates perception errors. 

	\item Contrast 
    \begin{equation}
        Pr(|C_D - C_t|_{t=t_k-t_{k-M}}^{t_k}) < P_{C}
    \end{equation}
    
    where $C_t$ is the contrast of the bounding box at time $t$, $C_D$ is the desired contrast from the training phase, and $P_C$ is the probabilistic threshold for contrast.  Highly varying contrast also modifies the details of the same object information over time.  Contrast consistency is one of the factors that we need to keep for more robust perception.  

	\item Entropy
    \begin{equation}
        Pr(|E_D - E_t|_{t=t_k-t_{k-M}}^{t_k}) < P_{E}
    \end{equation}
    where $E_t$ is the entropy of the bounding box at time $t$ and $E_D$ is the desired entropy from the training phase.  And $P_E$ is the probabilistic threshold for entropy.  Too blurred images or too sharpened images ruin even the state of the art deep-learning based object detection methods.  So, the entropy deviates from the desired value, it is highly possible that we have perception errors.  

\end{itemize}

Then the corresponding optimization formula to control contrast $(c_i (t)+\Delta c)$ with the cost function $J(c, \Delta c)$ is defined as, 

\begin{equation}
    \arg\min_{\Delta c} J(c, \Delta c) = \arg\min_{\Delta c} \sum_{i=1}^{n} |c_i(t) - c_D - \Delta c |
    \label{eq:ContrastOptimization}
\end{equation}

where  $c_i (t)$ is the contrast value of the $i^{th}$ detected object at time $t$, $c_D$ is the desired contrast value from the procedure of finding the probabilistic distributions of the probes, and $\Delta c$ is the system control input for contrast (which is the same as estimated deviation to apply to the perception module).   

For contrast control, once the desirable contrast deviation is acquired, we set up the expansion of histogram ranges to achieve that contrast changes using the peak-to-peak contrast (Michelson contrast) \cite{c10}.   The peak-to-peak contrast is defined in the following way:

\begin{equation}
    C(t) = \frac{I_{max} - I_{min}}{I_{max} + I_{min}} 
    \label{eq:MichelsonContrast}
\end{equation}

where $I_{max}$ is the maximum image intensity value and $I_{min}$ is the minimum image intensity value.  From this definition, we can expect a new contrast is supposed to be:

\begin{equation}
    C(t) = \frac{I_{max} - I_{min} + 2B}{I_{max} + I_{min}} 
    \label{eq:MichelsonContrastBound}
\end{equation}

where $B$ is the expanded histogram range to achieve the new contrast.  Since $\Delta c = C_{desired}-C(k)$, the histogram changing range changing amount will be

\begin{equation}
    B = \frac{\Delta c \cdot (I_{max} + I_{min})}{2}  
    \label{eq:ContrastBoundary}
\end{equation}

\section{Experiments}
\label{sec:experiments} 

\begin{figure}[ht]
  \centering
    \includegraphics[width=5in]{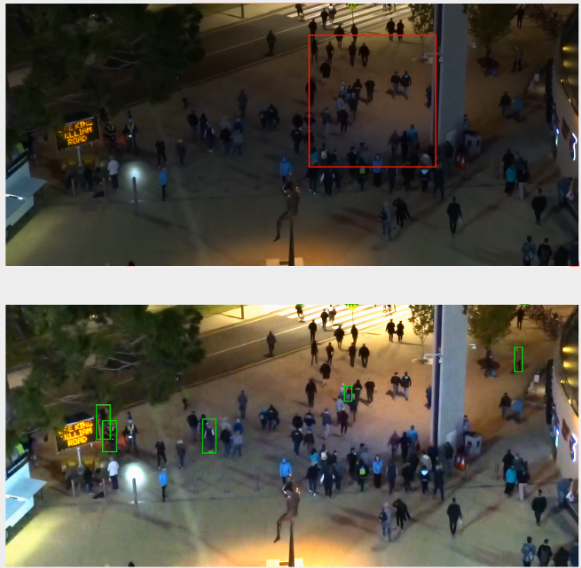}
  \caption{Sample of error reduction and detection improvement. On top are unique detections from a baseline YOLOv3. Below are unique detections using our proposed CogSense method. We intentionally remove common detections from both to increase readability.}
  \label{fig:SampleResult}
\end{figure}

In this section, we present test results with our contrast-based PSTL perception adaptation.  First, we present a test result on a video from the Multiple Object Tracking(MOT) Benchmark dataset \cite{c11}. Through our proposed method, we improve detection results as shown in Figure \ref{fig:SampleResult}. The red box in the upper figure is a erroneous person detection from the original image, and the green boxes in the lower figure are correct person detection newly added from the CogSense system.

\begin{figure}[ht]
  \centering
    \includegraphics[width=4in]{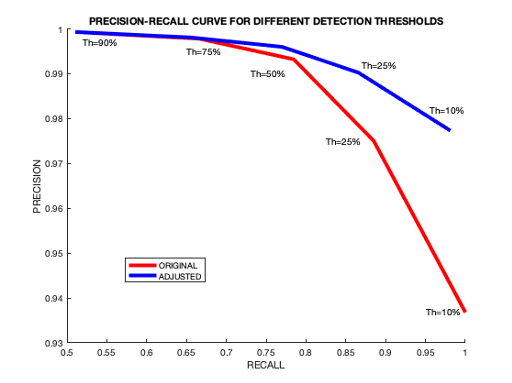}
  \caption{Performance Comparison - Precision vs Recall}
  \label{fig:PrecisionRecall}
\end{figure}

\begin{figure}[ht]
  \centering
    \includegraphics[width=4in]{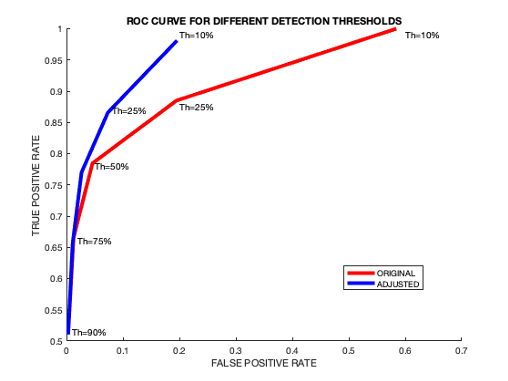}
  \caption{Performance Comparison - ROC curve}
  \label{fig:ROC}
\end{figure}

For supporting the proposed method with quantified performances, we also provide a precision-recall graph and a ROC curve in Figs. \ref{fig:PrecisionRecall} and \ref{fig:ROC}, respectively.  As shown in the graphs, the newly proposed approach presents better recall rates with the same precision, and also less false positive rates while achieving the same true positive rates.  Especially, in the ROC curve, with 10\% detection confidence thresholding, the false positive rate is reduced by 41.48\%.   

In addition to the above, we also tested more challenging video clips collected from \cite{c12,c13,c14,c15,c16}.  Then we compare the detection results from the original video sequences with those from the conventional Contrast Limited Adaptive Histogram Equalization (CLAHE) \cite{c5} and those from our proposed CogSense method.  Tables \ref{tbl:results_tp} and \ref{tbl:results_fp} show the true positive rates (the number of true positives over the sum of true positive and false positive detections) and false positive counts, respectively.  As shown in Table \ref{tbl:results_tp}, although the improvement rates are not that huge, the true positive rates are actually improved overall compared to the original method and the CLAHE method.  And if we look into the false positive counts in Table \ref{tbl:results_fp}, false positive counts with the proposed method are reduced by $12.5 \%$ comparing to the original video processing, and by $34.75 \%$ comparing to the CLAHE-applied video processing. The reason of CLAHE-applied videos' high false positive rates is that CLAHE-based image enhancement improves visual aspects to human, not for object detection.  

\begin{table}[ht]
 \caption{True positive rates from the original, CLAHE and the proposed CogSense approach}
    \begin{center}
    \begin{tabular}{ |c|c|c|c| } 
    \hline
    Video & Original & CLAHE & CogSense \\ 
    \hline
    Clip 1 \cite{c12} & 0.9804 & 0.9602 & \textbf{0.9817} \\
    \hline
    Clip 2 \cite{c13} & 0.9931 & \textbf{0.9931} & 0.9904 \\ 
    \hline
    Clip 3 \cite{c14} & 0.9829 & \textbf{0.9860} & 0.9835 \\
    \hline
    Clip 4 \cite{c14} & 0.9964 & 0.9977 & \textbf{0.9993} \\
    \hline
    Clip 5 \cite{c15} & 0.9822 & 0.9796 & \textbf{0.9834} \\
    \hline
    Clip 6 \cite{c16} & 0.9495 & 0.9065 & \textbf{0.9527} \\
    \hline
    \end{tabular}
    \label{tbl:results_tp}
 \end{center}
\end{table}

\begin{table}[ht]
 \caption{False positive counts from the original, CLAHE and the proposed CogSense approach}
    \begin{center}
    \begin{tabular}{ |c|c|c|c| } 
    \hline
    Video & Original & CLAHE & CogSense \\ 
    \hline
    Clip 1 \cite{c12} & 22 & 43 & \textbf{20} \\
    \hline
    Clip 2 \cite{c13} & 14 & \textbf{8} & 12 \\
    \hline
    Clip 3 \cite{c14} & 13 & \textbf{11} & 13 \\
    \hline
    Clip 4 \cite{c14} & 5 & 3 & \textbf{1} \\
    \hline
    Clip 5 \cite{c15} & 17 & 19 & \textbf{16} \\ 
    \hline
    Clip 6 \cite{c16} & 17 & 34 & \textbf{15} \\
    \hline
    \end{tabular}
    \label{tbl:results_fp}
 \end{center}
\end{table}

\begin{figure}[ht]
  \centering
  	\includegraphics[width=7in]{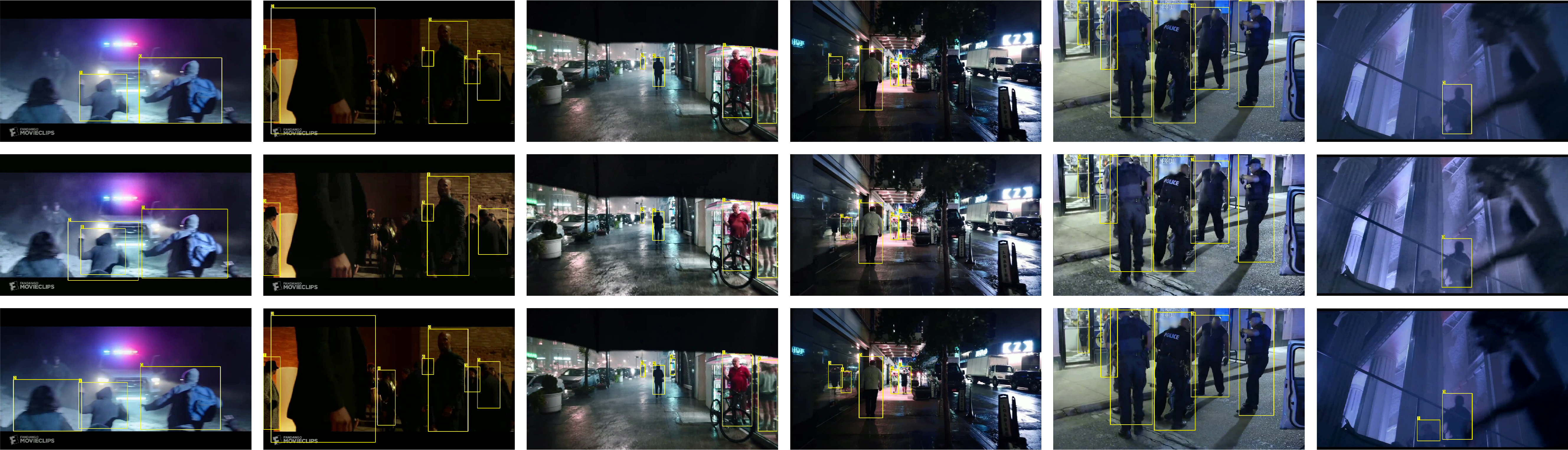}
  \caption{Qualitative results of our proposed CogSense method. From left to right, Clip 1, Clip 2, Clip 3, Clip 4, Clip 5, and Clip 6. From top to bottom: Original, CLAHE, and CogSense.}
  \label{fig:result_all}
\end{figure}

Figure \ref{fig:result_all} shows sample image triples from the three methods.  In the figure, each row corresponds to the individual clip number in order.  In Clip 1, the original video processing missed one person and the CLAHE-applied video processing still missed the same person, and even provided multiple detections from the same person.  On the other hand, our proposed method detects all three objects.  Also in Clip 2 and Clip 6, the proposed method detects more true positives.   And finally, the result of Clip 4 shows two different detections from the original video processing and CLAHE-applied video processing, however, our proposed method detected them both. 

As shown through the test results, our proposed CogSense method using PSTL-based constraints and detected-bounding-box-based optimization provides more robust detection outputs.

\section{Conclusion and Discussion}
\label{sec:conclusion} 

This paper presents CogSense, a new perception error detection and perception parameter adaptation method using the probabilistic signal temporal logic, and a contrast-based perception adaption approach as a specific case. The proposed method evaluates perception errors using heterogeneous probes of the detected objects and subsequently correct perception errors by solving a contrast-based optimization problem and other objected detection-based constraints generated from the probabilistic signal temporal logic system.  Our proposed contrast-based perception adaptation uses information only from the detected bounding boxes, and specifically our approach applies adaption to these boxes to improve object detection rather than unnecessarily applying adaptation to the entire image enhancement, which could have unintended consequences to other detections. In future work, we will increase the heterogeneity of probes to extend to other domains such as semantics and context, and pursue multi-modal perception parameter adaption by optimizing parameters to simultaneously control entities that extend beyond the current paper's demonstration of image based contrast to enhance the CogSense system's perception correction capabilities.


\addtolength{\textheight}{0cm} 







\end{document}